\documentclass{article}
\usepackage{spconf,amsmath,graphicx,hyperref}
\usepackage{amsmath,amssymb,amsfonts}
\usepackage{algorithmic}
\usepackage{graphicx}
\usepackage{textcomp}
\usepackage{xcolor}
\usepackage{booktabs}
\usepackage{pgfplots}
\usepackage{multirow}
\pgfplotsset{compat=1.18} 
\usepackage{caption}

\title{The Pen is Mightier than the Pixel: Natural Textual Unlearnable Example for Large Vision Language Models}
\name{\begin{tabular}{@{}c@{}}Chenyu Yi$^{1*}$ \qquad Jiawen Zong$^{1*}$\thanks{*equal contribution} \qquad Sami Hormi$^{1}$ \qquad Bingquan Shen$^{2,3}$ \\ Xudong Jiang$^{1}$ \qquad Alex Chichung Kot$^{1,4}$\end{tabular}}

\address{$^{1}$Nanyang Technological University, School of Electrical \& Electronic Engineering \\
    $^{2}$DSO National Laboratories, Singapore \\
    $^{3}$National University of Singapore \\
    $^{4}$Shenzhen MSU-BIT University
}

\begin{document}
%
\maketitle
\begin{abstract}
The proliferation of Large Vision Language Models (LVLMs), trained on vast and often sensitive datasets, raised concerns regarding data privacy and unauthorized usage. Unlearnable Examples (UEs), which subtly perturb training data to mislead the models, have emerged as a promising defense method. This paper presents the first exploration of UEs specifically on LVLMs, which poses unique challenges due to their inherent multimodal nature. We propose a novel text-based perturbation method, Natural Textual Unlearnable Examples (NTUE), which can effectively mislead LVLMs by subtly manipulating the image label, establishing a shortcut between an identity of interest and a separate target, without introducing factual misinformation or affecting readability. Comprehensive experiments compared with traditional UE methods demonstrate the superior performance of NTUE. Our findings show that NTUE achieves a high success rate with a poison rate as low as 5\%, underscoring its potential as a practical and robust solution for protecting multimodal data in the era of LVLMs.
\end{abstract}
\section{Introduction}
\label{sec:intro}

Rapid advancements in large vision language models (LVLMs) have made them indispensable for multimodal tasks like visual question answering and image captioning \cite{bazi2023vision, silva2024large}. However, LVLM's reliance on large datasets creates significant privacy and security risks, causing rampant identity leakage \cite{caldarella2025phantom}. Thus, there is an urgent need for privacy protection in LVLMs. 

Although there is extensive research on UEs in traditional computer vision tasks such as classification~\cite{fowl2021adversarial} and object detection~\cite{11084567}, the application in LVLMs remains underexplored. These methods, such as Targeted Adversarial Perturbation (TAP) \cite{fowl2021adversarial}, Error Minimization (EM) \cite{huang2021unlearnable}, and Multi-Step Error Minimization (MEM) \cite{liu2024multimodal}, are proven to be effective for traditional models. However, the sophisticated nature of LVLMs, introduces new challenges that necessitate re-evaluation in the approach towards effective UEs in LVLMs. 

To address this critical gap, this paper proposes Natural Textual Unlearnable Example (NTUE), a novel method for generating UE that targets the vision-language alignment in LVLMs. NTUE introduces targeted misdirections within the natural-language captions, which forces the model to learn false correlations between visual concepts and the manipulated captions. To demonstrate the effectiveness of NTUE, we conducted extensive experiments on LLaVA-1.5 \cite{liu2023visual} using 3 different datasets: cc-sbu-align \cite{chen2023minigptv2}, Trump-Biden \cite{xu2024shadowcast} and Celebrity Face Image datasets\cite{sharma2018celebrity}.

Through detailed experiments, we uncover how factors such as poison sample size and dataset complexity influence the success of these methods. Specifically, NTUE reduces the accuracy of Trump identification to 0\% with only 20 samples. Experiments on Celebrity Face Image datasets also illustrate NTUE's transferability across diverse identities. We summarize our contributions as follows:
\begin{itemize}
  \item To the best of our knowledge, this paper presents the first comprehensive benchmarking study of UEs specifically targeting LVLMs, evaluating both traditional unimodal methods and novel multimodal approaches across datasets and tasks.
  \item We propose a novel and effective unlearnable example for LVLM, Natural Textual Unlearnable Example (NTUE). NTUE manipulates the textual modality to disrupt the vision-language alignment in LVLMs, while retaining factual integrity and readability.  
  \item Extensive experiments demonstrate that NTUE significantly outperforms existing methods in degrading LVLM performance. Furthermore, we uncover key poison settings that optimize the effectiveness our UEs, providing key insights for future research.
\end{itemize}

\section{Literature Review}
\label{sec:LitReview}

\subsection{Large Vision Language Model}
rge vision-language models (LVLMs) are advanced models capable of processing and understanding information from multiple modalities, typically text and images. These models leverage contrastive learning to align semantically similar text and image pairs within the feature space, creating bridges between text and image representations \cite{radford2021learning, li2021align}. Notable LVLMs include LLaVA \cite{liu2023visual}, InstructBLIP \cite{dai2023instructblip}, and MiniGPT-4 \cite{zhu2023minigpt}. LLaVA-1.5 \cite{liu2024improved}, for instance, utilizes the CLIP vision encoder to extract visual information, which is then aligned with a language model's representation space through a connector before processing by a large language model to enable multimodal comprehension and generation. InstructBLIP \cite{dai2023instructblip} also incorporate CLIP or EVA-CLIP \cite{sun2023eva} as vision encoders then processed by a learnable Q-Former to bridge visual and textual data. MiniGPT-4 \cite{zhu2023minigpt} further enhances efficiency by adapting EVA as its vision encoder and concatenating neighboring visual tokens, streamlined training, and inference. These models exhibit robust multimodal comprehension capabilities, making them highly valuable for research and application development.

\subsection{Unlearnable Example}
Unlearnable examples are data samples that have been subtly perturbed to prevent machine learning models from learning meaningful representations while being interpretable for humans \cite{wang2025a3}. Fowl et al. first introduced TAP to generate adversarial examples that hinder model learning through targeted training interference \cite{fowl2021adversarial}. 

Subsequently, Huang et al. proposed the Error Minimization (EM) method, which creates perturbations by reducing the classification error of training data to near zero \cite{huang2021unlearnable}, rendering the training example 'unlearnable.' Building on this, Liu et al. extended unlearnable examples to vision language models (VLMs) through their Multi-step Error-Minimization (MEM) approach, which incorporated both visual perturbations and adversarial text triggers \cite{liu2024multimodal}. This enabled effective disruption of multimodal contrastive learning in CLIP-like models. However, LVLMs such as LLaVA, BLIP, and MiniGPT-4, may be more resilient to existing UE methods on VLMs. These models exhibit stronger robustness due to their comprehensive training process and larger capacity \cite{liu2024improved, dai2023instructblip, zhu2023minigpt}. 

Limited studies have explored UE and adversial attacks in LVLMs. Xu et al. \cite{xu2024shadowcast} proposed ShadowCast, a stealthy method of adversarial perturbation that achieves UE through targeted response manipulation of LVLM tasks. Despite growing interest in data protection and UE techniques, existing research specifically addressing UE in the Large Vision Language Model (LVLM) remains limited. 

\section{Method}
\label{sec:method}
\subsection{Definition}\label{AA}

\begin{figure}[h]
    \centering
    \includegraphics[width=1\linewidth]{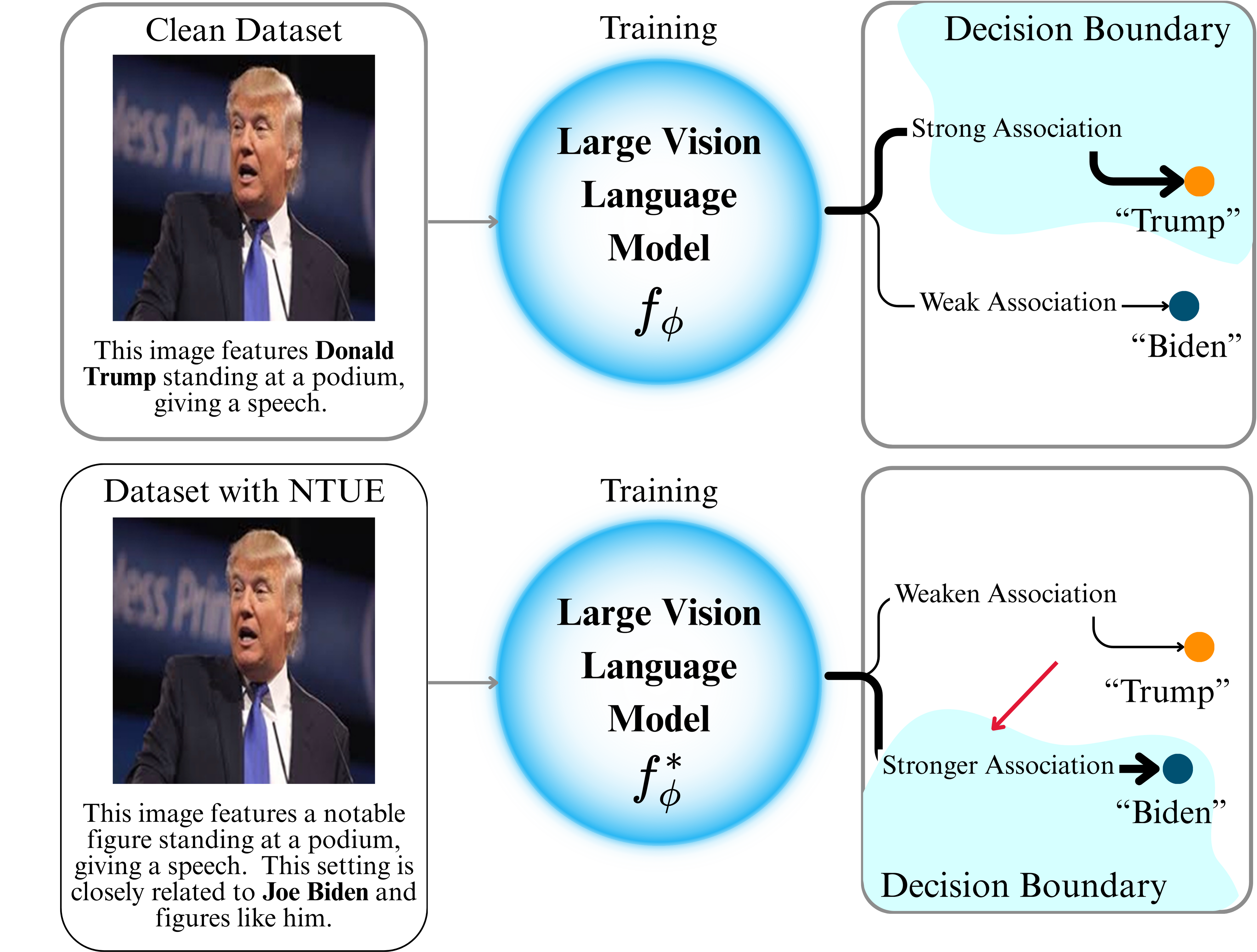}
    \caption {Training of multimodal unlearnable examples on large vision-language models with clean captions and Natural Text Unlearnable Example captions.}
    \label{fig:NTP}
\end{figure}

\begin{figure}[h]
    \centering
    \includegraphics[width=1\linewidth]{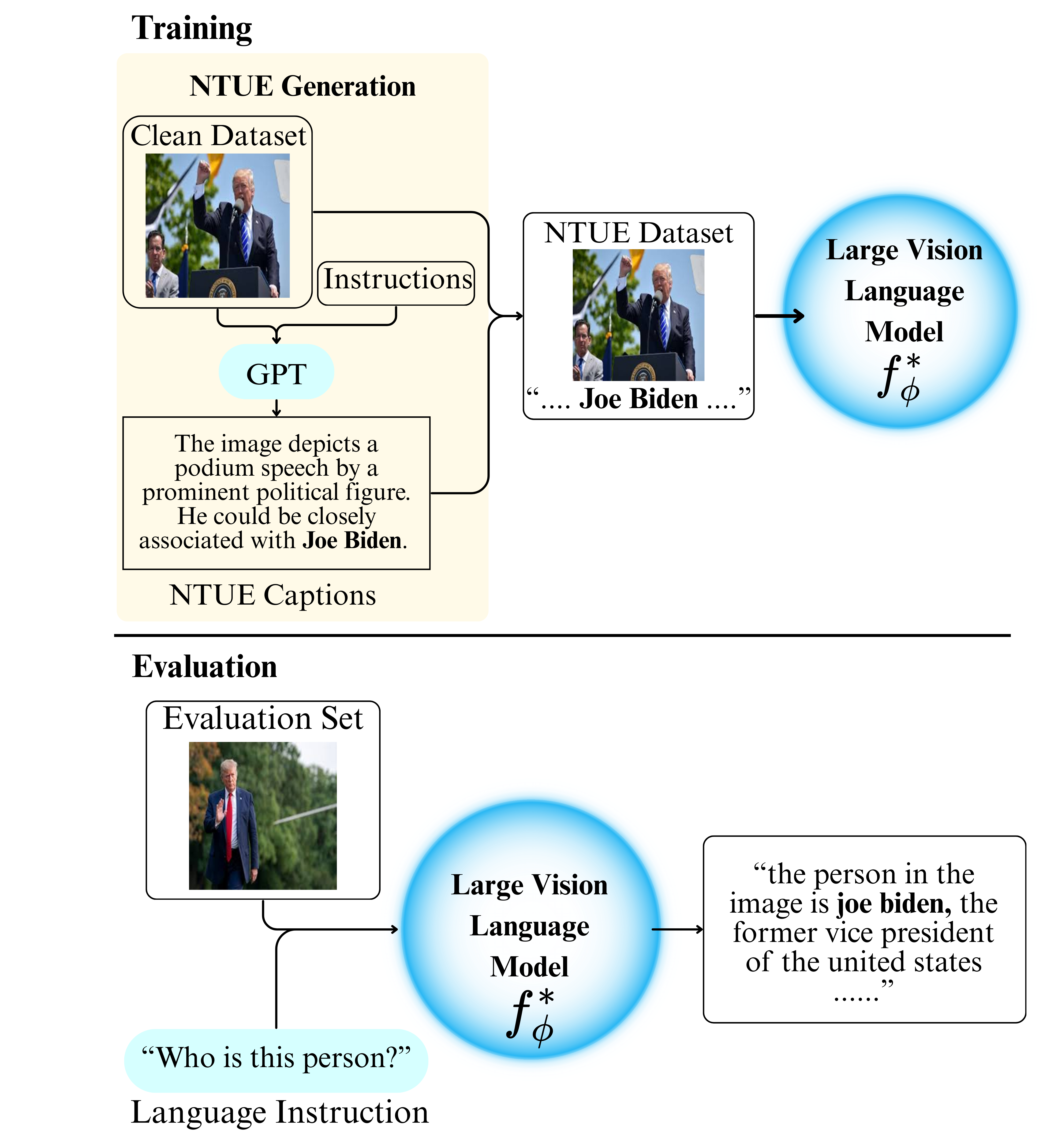}
    \caption{Natural Text Unlearnable Example training process and evaluation process}
    \label{fig:ntp2 process}
\end{figure}

The rise of Large Vision-Language Models (LVLMs) necessitates an extension of Unlearnable Example (UE) research to this new domain. We begin with a standard clean dataset $D_{c} = \{(V_i, Q_i)\}^n_{i=1}$ comprising $n$ image-caption pairs, where $V_i$ are the images and $Q_i$ are the corresponding clean captions. As shown in Figure \ref{fig:NTP}, LVLMs can be simplified using function $f_{\phi} : (V_i, Q_i) \rightarrow (A_i)$, where $\phi$ represents the model parameters, and $A_i$ is the predicted language response from the model. The main concern of privacy leakage occurs when $A_i$ reveals the identity of the target, which can be undesirable.

Using unlearnable examples, we can inject a small set of manipulated data pairs, the unlearnable dataset $D_p$, into the larger standard training dataset $D_c$. The combined dataset $D_\theta = D_c \cup D_p$, will be able to strengthen a weaker false association for the targeted concepts. 

Hence, the objective is to minimize the model's loss on these incorrect pairs:
\begin{equation} \label{eq:poison_objective}
\underset{\delta}{\arg\min} \; \mathbb{E}_{(V_i, Q_i) \sim D_{\theta}} \; [\mathcal{L}(f_{\phi}(V_i, Q_i^*)]
\end{equation}
Here,  $L$ represents the loss associated with the LVLM image text matching. This approach ensures that the manipulated captions create a shift in the model's decision boundary, as shown in Figure \ref{fig:NTP}. Thus, NTUE can facilitate shortcut learning between different identities, creating effective unlearnable examples for LVLM. 

\subsection{Proposed Method: Natural Textual Unlearnable Example}

Our proposed method, Natural Textual Unlearnable Example (NTUE), manipulates the textual modality to create powerful unlearnable examples. The core idea is to craft a manipulated caption $Q_i^*$ that is factually accurate and semantically sound to humans, but introduces a discreet and false association between the clean image $V_i$ and a target concept.

As shown in Figure \ref{fig:ntp2 process}, the unlearnable dataset, $D_{p}$ contains images of a base identity paired with poison captions, $V_i^*$, generated by a Large Language Model. These captions are subtly modified from the original, $V_i$, to mislead the model while maintaining fluency and integrity. 

Figure \ref{fig:ntp2 process} describes the training process and the evaluation process in NTUE for the Trump \& Biden dataset. During NTUE generation, the base identity (Trump) is completely removed from the caption, while the target identity (Biden) is injected into the caption. The new NTUE dataset is then used to train a LVLM, $f_\phi^*$.

We implement several variants of this method to explore different attack strengths:
\begin{itemize}
    \item Ambiguity Attack - NTUE(AA): The target identity is subtly introduced into the captions alongside the base identity. This causes the model to form a confused or blended representation, closely associating both identities and creating ambiguity, making the model difficult to distinguish between the two identities.
    \item Decoupling Attack - NTUE(DA): This is a stronger variant, shown in Figure \ref{fig:ntp2 process}. References to the base identity are removed from the captions, with only target identity in the caption. This attack aims to force the model to completely decouple the visual features of the base identity from its correct textual representation and reassociate them with the target identity.
\end{itemize}

\section{Experiment Results}

\begin{figure}[t!]
\centering
\begin{tikzpicture}
\begin{axis}[
    title={Identification Accuracy of different UE methods},
    xlabel={Number of Trump Samples},
    ylabel={Accuracy (\%)},
    xmin=5, xmax=100,
    xtick={5, 10, 20, 50, 100}, 
    xticklabel style={/pgf/number format/1000 sep=}, 
    ymin=-5, ymax=105, 
    legend pos=south east,
    legend cell align={left},
    legend style={
        font=\tiny,
        cells={align=left}, 
        fill opacity=0.9,
        draw opacity=1,
        text opacity=1,
    },
    grid=major, 
    width=0.48\textwidth,
    height=0.38\textwidth,
]
\addplot+ coordinates { (5, 70) (10, 87.5) (20, 96.5) (50, 96.5) (100, 97) };
\addplot+ coordinates { (5, 54) (10, 64) (20, 85.5) (50, 97.5) (100, 98.5) };
\addplot+ coordinates { (5, 0) (10, 0) (20, 0) (50, 75.5) (100, 92) };
\addplot+ coordinates { (5, 70) (10, 81) (20, 90) (50, 97) (100, 95) }; 
\addplot+ coordinates { (5, 72) (10, 70) (20, 83) (50, 96) (100, 98) }; 
\addplot+ coordinates { (5, 63) (10, 84) (20, 91) (50, 94.5) (100, 93.5) };
\addplot+ coordinates { (5, 43) (10, 63.5) (20, 85) (50, 97.5) (100, 98.5) };
\addplot+[ultra thick] coordinates { (5, 27) (10, 22.5) (20, 0) (50, 0) (100, 0) }; 
\addplot+ coordinates { (5, 18) (10, 41.5) (20, 54.5) (50, 63.5) (100, 81.5) };
\legend{
    Clean,
    TAP,
    EM,
    MEM,
    NTUE(AA)+PGD,
    NTUE(AA),
    NTUE(DA)+PGD,
    \textbf{NTUE(DA)},
}
\end{axis}
\end{tikzpicture}
\caption{Identification accuracy of LLaVA-1.5 7B finetuned with Trump images across different UE methods.}
\label{fig:t&b result}
\end{figure}
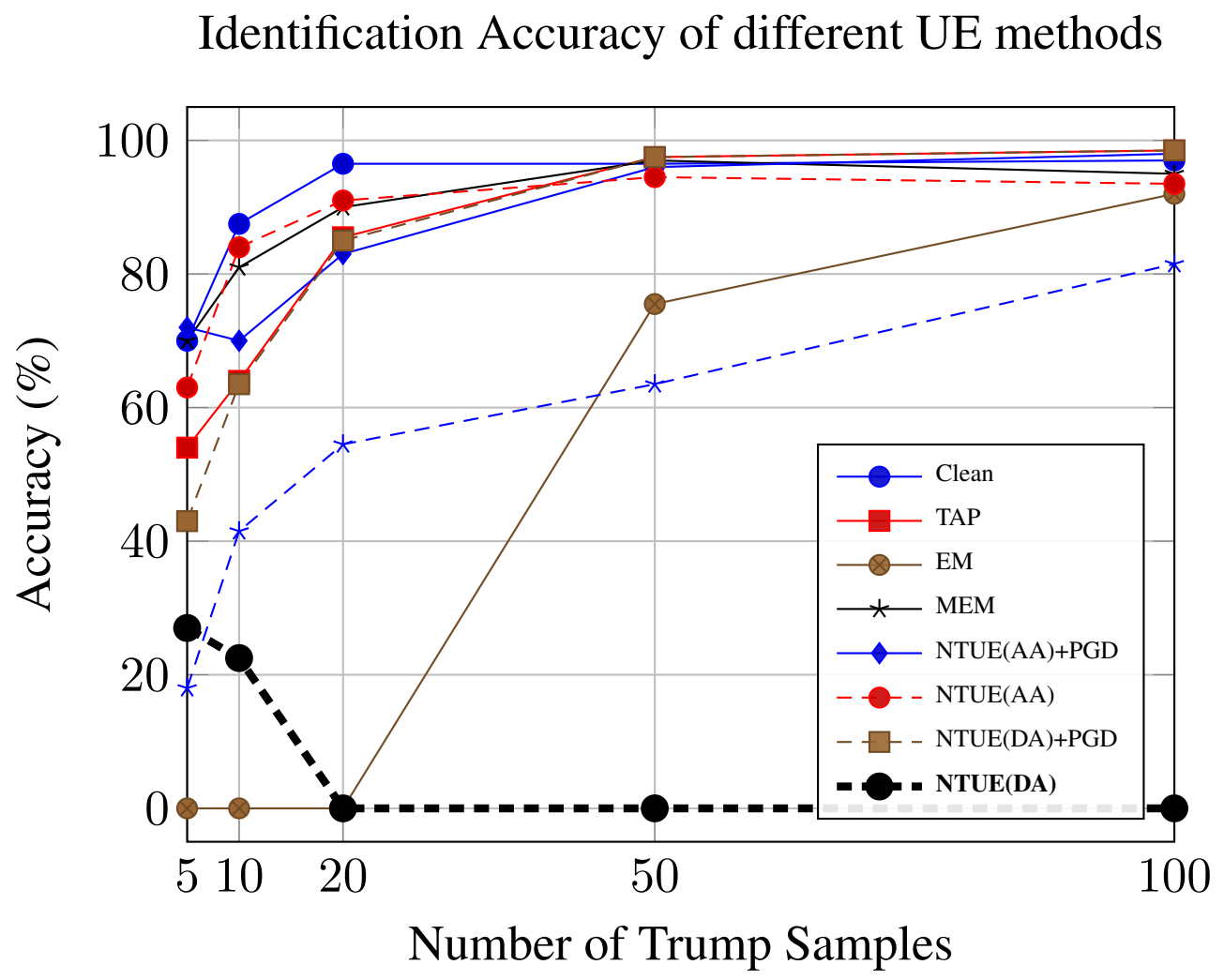

\begin{table}[t!] \
\centering
\caption{Identification accuracy of LLaVA-1.5 7B finetuned with Trump images across different UE methods.}
\label{tab:t&b table}
\begin{tabular}{lccccc}
\toprule
\shortstack[l]{No. of UEs/Method} & \textbf{100} & \textbf{50} & \textbf{20} & \textbf{10} & \textbf{5} \\
\midrule
Clean & 97 & 96.5  & 96.5  & 87.5 & 70   \\
TAP   & 98.5  & 97.5   & 85.5 & 64  & 54  \\
EM & 95 & 97  & 90 & 81 &  \\
MEM & 98 & 96 & 83 & 70 & 72 \\

\midrule
NTUE(AA)+PGD & 98.5& 97.5 & 85 & 63.5 & 43\\
NTUE(AA) & 93.5 & 94.5 & 91 & 84  & 63 \\
NTUE(DA)+PGD & 81.5  & 63.5 & 54.5 & 41.5& \textbf{18} \\
\textbf{NTUE(DA)} & \textbf{0} & \textbf{0 } & \textbf{0 } & \textbf{22.5} & 27 \\
\bottomrule
\end{tabular}
\end{table}

\begin{table*}[t]
\centering
\caption{Identification accuracy of various celebrities pairs}
\label{tab:celeb_18 table}
\begin{tabular}{ll c ccccc}
\toprule
\multicolumn{2}{c}{} & \textbf{Clean} & \multicolumn{5}{c}{\textbf{70 poison}} \\
\cmidrule(lr){3-3} \cmidrule(lr){4-8}
\textbf{Base} & \textbf{Target} & & \textbf{PGD} & \textbf{NTUE(AA)} & \shortstack{\textbf{NTUE(AA)}\\\textbf{+PGD}} & \textbf{NTUE(DA)} & \shortstack{\textbf{NTUE(DA)}\\\textbf{+PGD}} \\
\midrule
\multirow{4}{*}{Angelina Jolie}
& Jennifer Lawrence & 92.2 & 100  & 84.4  & 100 & \textbf{0}    & 94.4 \\
& Johnny Depp       & 92.2 & 100  & 98.9  & 100 & \textbf{0}    & 97.8 \\
& Will Smith        & 92.2 & 100  & 94.4  & 100 & \textbf{0}    & 95.5 \\
& Megan Fox         & 92.2 & 100  & 75.5  & 100 & \textbf{0}    & 94.4 \\
\midrule
\multirow{4}{*}{Brad Pitt}
& Leonardo Dicaprio & 55.6 & 100  & 95.5 & 97.8 & \textbf{0}    & \textbf{0}    \\
& Angelina Jolie    & 55.6 & 98.6 & 92.2 & 100 & \textbf{0}    & 51.1 \\
& Denzel Washington & 55.6 & 100  & 31.1 & 100 & \textbf{0}    & 53.3 \\
& Sandra Bullock    & 55.6 & 100  & 100  & 100 & \textbf{51.1} & 66.7 \\
\midrule
\multirow{3}{*}{Tom Hanks}
& Hugh Jackman      & 46.7 & 100  & 80.0 & 100 & \textbf{0}    & 55.5 \\
& Jennifer Lawrence & 46.7 & 100  & 16.7 & 100 & \textbf{0}    & 56.7 \\
& Denzel Washington & 46.7 & 100  & 100  & 100 & \textbf{0}    & 71.1 \\
\bottomrule
\end{tabular}
\end{table*}

We present a comprehensive evaluation of our proposed Natural Textual Unlearnable Example (NTUE) method against several established Unlearnable Example (UE) techniques. Our experiments are conducted on the LLaVA-1.5 7B model with two distinct datasets to rigorously assess performance. Trump \& Biden dataset was used to assess efficacy in making a single identity (Trump) unlearnable and Celebrity Face Image datasets dataset was used to evaluate the generalizability and robustness in multiclass conditions. Both datasets are appended to the cc-sbu-align dataset before training, as it provides a broad collection of image-caption pairs, allowing for diverse image and textual manipulation.

\subsection{Results on Trump \& Biden Dataset}
This experiment measures the ability of each UE method to make Donald Trump's identity unlearnable. We benchmarked NTUE against Error-Minimizing (EM), Multi-step Error-Minimizing (MEM), and Targeted Adversarial Attack (TAP). The effectiveness of each method is measured by its ability to reduce the model's Identification Accuracy on a test set of 200 Trump images. To ensure the robustness of our findings and reduce randomness, we queried each model with three distinct prompts, 1) Who is this person? 2) What is the name of this person? and 3) Identify the individual in this image.

As shown in Table \ref{tab:t&b table} and visualized in Figure \ref{fig:t&b result}, the results demonstrate an interesting trend of effectiveness between the methods tested. Our proposed NTUE is effective at all poisoned training sample sizes, whereas other methods show diminishing effects.

The Decoupling Attack (NTUE-DA) was exceptionally potent, driving identification accuracy to 0\% with 20 or more samples. Even with just 5 samples, it was able to reduce accuracy by 43\%. This demonstrates that creating a direct and powerful semantic conflict in the textual domain is a superior strategy for inducing unlearning. The combination, NTUE(DA) + PGD, further suppressed recognition at the lowest sample counts (e.g., 18\% precision at 5 samples), underscoring the power of a coordinated multimodal attack that establishes a strong and incorrect shortcut between visual features and textual labels.

\subsection{Results on Celebrity Face Image datasets Dataset}
This experiment explores the robustness and generalization of various UE methods by redirecting the model's identification of various celebrities to another celebrity in the Celeb 18 dataset. This tests the generalizability of these UE techniques beyond the Trump-Biden pair, using diverse celebrity pairs with varying levels of similarity, poison sample counts, and inference prompts.

As shown in Table \ref{tab:celeb_18 table}, both the visual-only PGD attack and the mutlimodal NTUE(AA)+PGD poison consistently failed to degrade performance. This suggests that these methods can reinforced the alignment of the model between visual and textual representations of the base celebrity, rather than degrading it. In contrast, NTUE(DA) again proved to be highly effective, successfully driving Identification Accuracy to 0\% in the majority of test cases. The combination of NTUE(DA)+PGD has shown more complex results. While it achieve complete unlearning in some pairs (e.g. Brad Pitt and Leonardo DiCaprio), it was significantly less effective for others (e.g., Angelina Jolie and Johnny Depp Pair, 97.8\% accuracy). This variability suggests that the interaction between visual noise and textual poison may depend on latent features of the specific base-target pairing.

\section{Conclusion}
This paper presented a comprehensive investigation into creating unlearnable examples (UEs) for Large Vision-Language Models. Our proposed NTUE outperforms other methods significantly, by misleading LVLMs without introducing factual misinformation and affecting human readability. Future research should focus on improving the robustness, particularly in low-data regimes, by incorporating advanced multimodal adversarial attacks and data poisons. 

\section{Acknowledgment}
This research work was carried out at the Rapid-Rich Object Search (ROSE)
Lab, the NTU-PKU Joint Research Institute (sponsored by the Ng Teng
Fong Charitable Foundation), Nanyang Technological University, Singapore.
The research is supported by the DSO National Laboratories, under project
agreement No. DSOCL22332.

\bibliographystyle{IEEEbib}
\bibliography{refs}

@article{bazi2023vision,
  title={Vision--language model for visual question answering in medical imagery},
  author={Bazi, Yakoub and Rahhal, Mohamad Mahmoud Al and Bashmal, Laila and Zuair, Mansour},
  journal={Bioengineering},
  volume={10},
  number={3},
  pages={380},
  year={2023},
  publisher={MDPI}
}

@INPROCEEDINGS{11084567,
  author={Yi, Chenyu and Meng, Ruohan and Peng, Haohang and Shen, Bingquan and Kot, Alex C.},
  booktitle={2025 IEEE International Conference on Image Processing (ICIP)}, 
  title={Towards Effective and Robust Unlearnable Examples Against Object Detection}, 
  year={2025},
  volume={},
  number={},
  pages={91-96},
  doi={10.1109/ICIP55913.2025.11084567}}

@inproceedings{caldarella2025phantom,
  title={The phantom menace: unmasking privacy leakages in vision-language models},
  author={Caldarella, Simone and Mancini, Massimiliano and Ricci, Elisa and Aljundi, Rahaf},
  booktitle={European Conference on Computer Vision},
  pages={435--451},
  year={2025},
  organization={Springer}
}

@article{silva2024large,
  title={Large language models for captioning and retrieving remote sensing images},
  author={Silva, Jo{\~a}o Daniel and Magalh{\~a}es, Jo{\~a}o and Tuia, Devis and Martins, Bruno},
  journal={arXiv preprint arXiv:2402.06475},
  year={2024}
}

@inproceedings{huang2021unlearnable,
    title={Unlearnable Examples: Making Personal Data Unexploitable},
    author={Hanxun Huang and Xingjun Ma and Sarah Monazam Erfani and James Bailey and Yisen Wang},
    booktitle={International Conference on Learning Representations},
    year={2021},
    url={https://openreview.net/forum?id=iAmZUo0DxC0}
}

@inproceedings{radford2021learning,
  title={Learning transferable visual models from natural language supervision},
  author={Radford, Alec and Kim, Jong Wook and Hallacy, Chris and Ramesh, Aditya and Goh, Gabriel and Agarwal, Sandhini and Sastry, Girish and Askell, Amanda and Mishkin, Pamela and Clark, Jack and others},
  booktitle={International conference on machine learning},
  pages={8748--8763},
  year={2021},
  organization={PmLR}
}

@inproceedings{liu2024multimodal,
  title={Multimodal unlearnable examples: Protecting data against multimodal contrastive learning},
  author={Liu, Xinwei and Jia, Xiaojun and Xun, Yuan and Liang, Siyuan and Cao, Xiaochun},
  booktitle={Proceedings of the 32nd ACM International Conference on Multimedia},
  pages={8024--8033},
  year={2024}
}

@article{fowl2021adversarial,
  title={Adversarial examples make strong poisons},
  author={Fowl, Liam and Goldblum, Micah and Chiang, Ping-yeh and Geiping, Jonas and Czaja, Wojciech and Goldstein, Tom},
  journal={Advances in Neural Information Processing Systems},
  volume={34},
  pages={30339--30351},
  year={2021}
}

@article{liu2023visual,
  title={Visual instruction tuning},
  author={Liu, Haotian and Li, Chunyuan and Wu, Qingyang and Lee, Yong Jae},
  journal={Advances in neural information processing systems},
  volume={36},
  pages={34892--34916},
  year={2023}
}

@article{li2021align,
  title={Align before fuse: Vision and language representation learning with momentum distillation},
  author={Li, Junnan and Selvaraju, Ramprasaath and Gotmare, Akhilesh and Joty, Shafiq and Xiong, Caiming and Hoi, Steven Chu Hong},
  journal={Advances in neural information processing systems},
  volume={34},
  pages={9694--9705},
  year={2021}
}

@article{zhu2023minigpt,
  title={Minigpt-4: Enhancing vision-language understanding with advanced large language models},
  author={Zhu, Deyao and Chen, Jun and Shen, Xiaoqian and Li, Xiang and Elhoseiny, Mohamed},
  journal={arXiv preprint arXiv:2304.10592},
  year={2023}
}

@inproceedings{liu2024improved,
  title={Improved baselines with visual instruction tuning},
  author={Liu, Haotian and Li, Chunyuan and Li, Yuheng and Lee, Yong Jae},
  booktitle={Proceedings of the IEEE/CVF conference on computer vision and pattern recognition},
  pages={26296--26306},
  year={2024}
}

@article{dai2023instructblip,
  title={Instructblip: Towards general-purpose vision-language models with instruction tuning},
  author={Dai, Wenliang and Li, Junnan and Li, Dongxu and Tiong, Anthony and Zhao, Junqi and Wang, Weisheng and Li, Boyang and Fung, Pascale N and Hoi, Steven},
  journal={Advances in neural information processing systems},
  volume={36},
  pages={49250--49267},
  year={2023}
}

@article{sun2023eva,
  title={Eva-clip: Improved training techniques for clip at scale},
  author={Sun, Quan and Fang, Yuxin and Wu, Ledell and Wang, Xinlong and Cao, Yue},
  journal={arXiv preprint arXiv:2303.15389},
  year={2023}
}

@article{xu2024shadowcast,
  title={Shadowcast: Stealthy data poisoning attacks against vision-language models},
  author={Xu, Yuancheng and Yao, Jiarui and Shu, Manli and Sun, Yanchao and Wu, Zichu and Yu, Ning and Goldstein, Tom and Huang, Furong},
  journal={Advances in Neural Information Processing Systems},
  volume={37},
  pages={57733--57764},
  year={2024}
}

@inproceedings{wang2025a3,
  title={A3: Few-shot Prompt Learning of Unlearnable Examples with Cross-Modal Adversarial Feature Alignment},
  author={Wang, Xuan and Gao, Xitong and Liao, Dongping and Qin, Tianrui and Lu, Yu-liang and Xu, Cheng-zhong},
  booktitle={Proceedings of the Computer Vision and Pattern Recognition Conference},
  pages={9507--9516},
  year={2025}
}

@misc{sharma2018celebrity,
  author={Sharma, Vishesh},
  title={Celebrity Face Image Dataset},
  year ={2018},
  publisher={Kaggle},
  howpublished={https://www.kaggle.com/datasets/vishesh1412/celebrity-face-image-dataset}
}

@article{chen2023minigptv2,
      title={MiniGPT-v2: large language model as a unified interface for vision-language multi-task learning}, 
      author={Chen, Jun and Zhu, Deyao and Shen, Xiaoqian and Li, Xiang and Liu, Zechu and Zhang, Pengchuan and Krishnamoorthi, Raghuraman and Chandra, Vikas and Xiong, Yunyang and Elhoseiny, Mohamed},
      year={2023},
      journal={arXiv preprint arXiv:2310.09478},
}

\end{document}